# Deep Learning Framework for Measuring the Digital Strategy of Companies from Earnings Calls

**Ahmed Ghanim Al-Ali[1,2], Robert Phaal[1], and Donald Sull[3]**
[1]Department of Engineering, University of Cambridge, Cambridge, UK
[2]DeepOpinion, Innsbruck, Austria
[3]Sloan School of Management, MIT, Boston, USA
ahmed.alali@deepopinion.ai, rp108@cam.ac.uk, dsull@mit.edu

## Abstract

Companies today are racing to leverage the latest digital technologies, such as artificial intelligence, blockchain, and cloud computing. However, many companies report that their strategies did not achieve the anticipated business results. This study is the first to apply state-of-the-art NLP models on unstructured data to understand the different clusters of digital strategy patterns that companies are Adopting. We achieve this by analyzing earnings calls from Fortune's Global 500 companies between 2015 and 2019. We use Transformer-based architecture for text classification which show a better understanding of the conversation context. We then investigate digital strategy patterns by applying clustering analysis. Our findings suggest that Fortune 500 companies use four distinct strategies which are product-led, customer experience-led, service-led, and efficiency-led[1]. This work provides an empirical baseline for companies and researchers to enhance our understanding of the field.

## 1 Introduction

The use of digital technologies is transforming the modern economy with significant implications for businesses. As a result, organizations are perceiving digital technologies to present both growth opportunities and existential threats (Sebastian et al., 2017). Despite efforts to adopt such technologies, research demonstrates that the success rate in improving business performance is very low due to the lack of a coherent digital strategy (Correani et al., 2020). A positive research step to address this challenge is enhancing the understanding of current approaches to digital strategy.

Corporate documents are increasingly being used to understand the performance of organizations for various purposes. Examples are, predicting expected returns from financial reports (Theil et al., 2019) and measuring compliance from sustainability reports (Smeuninx et al., 2020).

In the present study, we use earnings calls transcripts, as they provide rich insights into companies' activities, specifically on their digital strategy. We quantify the approach taken by companies based on the progress made on the following three components related to the digital strategy (Vial, 2019):

- **Business value:** the expressed value of using a specific digital solution (e.g., enhancing customer experience and increasing operational efficiency)
- **Strategy management**: the policies and practices in place to support the implementation of digital solutions (e.g., setup of innovation lab, acquisition of startups, and use of agile methods)
- **Digital technology:** the technology used as a part of the identified digital solution (e.g., artificial intelligence, Internet of things, and robotics)

In this work, we offer two contributions. First, we set the baseline for using deep learning based NLP models to measure the digital strategy of companies from earnings calls transcripts. Second, we apply this framework to investigate the digital strategy of Fortune's Global 500 companies.

---

[1] Experiment repository: `https://github.com/alali3030/earnings_calls_NLP`

## 2 Related Work

There have been a few recommendations for the digital strategy to be customer focused, product focused (Sebastian et al., 2017), or business model focused (Vial, 2019). However, an empirical investigation of digital strategy archetypes is still needed (Tekic & Koroteev, 2019).

A few NLP-driven publications have investigated the digital strategy of companies. This included network analytics of website tags (Stoehr et al., 2019), document clustering of companies' description (Riasanow et al., 2020), and keyword analysis of financial reports (Pramanik et al., 2019). While all these studies revealed insights into the digital strategy of companies, they were exploratory in nature with limited quantitative evaluation of digital strategy components.

Earnings calls transcripts are known to be rich data sources and considered to be more informative and insightful than company filing (Frankel et al., 1999). Moreover, they can be leveraged for text classification to identify decisions and activities that companies take (Keith & Stent, 2019). Therefore, we propose text classification of digital strategy related topics from earnings calls transcripts as detailed in Section 3.

## 3 Methodology

There is no generic framework to measure company performance, as it can be topic specific. Measuring digital strategy retrospectively is accomplished by identifying the progress made on its components, also known as digital maturity (Gurbaxani & Dunkle, 2019). From an NLP perspective, this requires identifying topics of interest in the text, followed by assigning a maturity score to it. The closest text classification task to this is aspect-based sentiment analysis, which is proven to be effective for mining aspects related to customers' opinions (Jiang et al., 2019). Therefore, we adapt this task to measure companies' digital strategy and refer to it as Aspect-based Maturity Analysis (ABMA).

In this case, we propose *Aspects* that refer to 17 coarse-grained topics from the three components of digital strategy presented in the introduction (business value [4 aspects], strategy management [2 aspects], and digital technology [11 aspects]). The design and selection on the 17 topics was based on extensive literature review of various publications on digital strategy components (Al-Ali, 2020). We find multi-label classification suitable as the labels are not mutually exclusive. *Maturity*, on the other hand, refers to the progress made with a given aspect over four discrete steps, including (1) plan, (2) pilot, (3) release, and (4) pioneer (EY & Microsoft, 2019). We propose the use of multi-class classification given that maturity runs on a discrete scale. Model labels are shown in Table 1. See Appendix B for the list of labels definitions. The following is an example of the classification process from the earnings calls dataset:

**TEXT:** *"We're putting most of our efforts right now—are continuing to—into our robotics program. We think it's been a great addition to our fulfillment capacity."*
{Aspects: [robotics, operations], Maturity: released (3)}

| Task | Labels |
|---|---|
| Aspect | **Business Value**: *digital products, digital customer experience, digital operations, digital business model* <br> **Strategy Management:** *strategy enablers, strategy practices* <br> **Digital technologies***: AI, analytics, Internet of things, blockchain, cloud computing, mobile, social media, robotics, augmented reality, virtual reality, 3D printing* |
| Maturity | *Plan, Pilot, Release, Pioneer* |

Table 1: Labels description for Aspect and Maturity classification.

## 4 Experiment Setup

The objective of the experiment is to demonstrate the utility of ABMA in measuring the digital strategy of Fortune 500 companies using earnings calls transcripts. The process included pre-processing and structuring the text, filtering irrelevant content, classifying the text, and aggregating the results to the company level. We then clustered the results in Section 5 to identify common digital strategy patterns.

### 4.1 Data and Pre-processing

We chose Fortune's Global 500 companies as a suitable sample based on their diversity and scale of digital activities (Fortune.com, 2019). Our data consists of 4,911 earnings calls transcripts for 304 companies covering the five years between January 1, 2015 and December 31, 2019. 195 companies were excluded due to missing or inconsistent data from the source. The data was then tabularized with features, including company name, ticker symbol, date of call, and call transcript.

Performing sentence-level splitting resulted in approximately 3.2 million sentences. We conducted a keyword search to select relevant sentences and maintain a dense dataset. The keywords included 275 domain-specific terms[3] from the 17 topics related to the three digital strategy components. This resulted in 46,277 sentences showing that around 1.46% of earning calls discussions are digital related. We also added previous and following sentences to capture sufficient context. We refer to each text block as a document.

### 4.2 Text Classification

Based on the proposed approach, we designed two-stage text classification architectures. The first model is multi-label to detect the occurrence of an aspect in a document, whereas the second model assigns a maturity class to it[4]. As an experiment, we hand-annotated 1,300 examples from a random sample on aspects and their respective maturity levels as shown by the illustrative example in Section 3. We split the labeled data to 80/10/10 between training, validation, and testing respectively.

Given the absence of a benchmark, we compared the performance of several text classification models, including transformer architecture. We trained the transformer models by applying a discriminative classification fine-tuning to maximize the utilization of the language model pre-training (Howard & Ruder, 2018). The training process also included gradual unfreeze of each layer with a slanted triangular learning rate to aid the convergence of the model parameters towards task-specific features (Howard & Ruder, 2018). We found that the pretrained RoBERTa (Liu et al., 2019) performed best without further language model fine-tuning, as shown in Table 2. The main difference was that RoBERTa was able to achieve higher accuracy on scarce labels. Moreover, we found RoBERTa to generalize better based on context with unseen terminologies. The qualitative evaluation of the output showed reasonable performance based on the training dataset size, in which errors were commonly attributed to imprecisely expressed sentences.

| Model | Aspect | | | Maturity | | |
|---|---|---|---|---|---|---|
| | Precision | Recall | F1-weighted | Precision | Recall | F1-weighted |
| RoBERTa $_{base}$ | **68.9** | **67.5** | **67.4** | **59.1** | **58.2** | **58.2** |
| BERT $_{base}$ | 65.5 | 63.3 | 62.0 | 56.8 | 56.0 | 55.7 |
| LSTM+GloVe | 66.4 | 60.2 | 62.5 | 51.3 | 50.9 | 50.6 |
| NB SVM | 52.1 | 46.0 | 47.0 | 49.0 | 51.3 | 49.0 |

Table 2: Aspect and Maturity experiment results.

### 4.3 Processing Output

Using the described text classification approach, we detected 61,872 aspect occurrences in 27,198 documents on 295 companies. This shows that 58.7% of the filtered dataset referred to a specific digital strategy-related activity by the company.

To aggregate results from the documents to the company level, we applied several transformation steps. First, we one-hot encoded all the aspects to obtain a binary feature vector. Second, we multiplied each vector of a document with its respective maturity class (1–4).

Differentiating companies that exhibit multiple maturity levels of an aspect in the same year was important. Taking a weighted average was penalizing companies for simultaneous maturity levels. Therefore, we treat the maturity of an aspect as a checklist, in which the score is calculated by summing each identified maturity class within the same year for a given aspect. We then calculated the mean across the five years, which resulted in a maturity score between 0 and 10 for each aspect. As a result, the dataset was a matrix of *295 companies × 18 features* (*17 aspects-maturity scores + mean maturity*).

---
[3] Appendix A provides a list of the keyword terms for each label.
[4] Appendix B outlines the classification labels definitions, annotation process, and label count in the dataset.

## 5   Clustering of Results

Companies may adopt various digital strategies based on sector, digital maturity, and business scope. To identify the common digital strategy archetypes, we clustered the data. As k-means clustering assumes an equal mean and variance, we standardized the data by applying *MinMax* scaling and *log transformation*. There were few sparse features, so we chose 12 dense features *(non-zero values > 40% of all companies)*. We also applied t-SNE[5] for dimensionality reduction, as it significantly improved clustering performance (Maaten & Hinton, 2008). Plotting the Silhouette score for clustering showed multiple peaks demonstrating inherent hierarchy in the data. Upon visual inspection, we found 10 clusters to be meaningful representation. The cluster map is illustrated in Figure 1 (a).

Investigating the cluster map revealed two main insights. First, clusters generally included companies in the same or a related sector. This indicates that the digital strategy can be sector specific. Second, we found a few companies from various sector clustered with predominantly digital native companies from the technology sector. Some examples are ABB and Siemens from the industrial sector are in cluster 2 due to their capabilities in automation and robotics technologies while the majority of their peer companies are in cluster 6. This indicates that some companies have made significant progress in digitally transforming their business.

We investigate the clusters by calculating the mean maturity value for each feature across the 10 clusters. We found four distinct digital strategies shared between eight clusters, whereas the remaining two had very limited digital maturity. The four digital strategy archetypes are product led *(cluster 2)*, customer experience led *(cluster 4)*, service led *(clusters 0,7,8,9)*, and efficiency led *(clusters 3,6)*, as illustrated by the radar charts in Figure 1 (b). We also found that while companies lead with a specific business value, such as customer experience, the vast majority demonstrate some level of progress across all areas. Therefore, focusing on a single area, as some authors argue, might not be practical due to the interdependencies between business functions (Sebastian et al., 2017; Westerman et al., 2014).

Our findings have two main implications. First, managers can use the identified archetypes as a baseline for digital strategy formulation and as a tool to benchmark the progress made against relevant companies. Second, researchers can build on our findings to investigate digital strategy further.

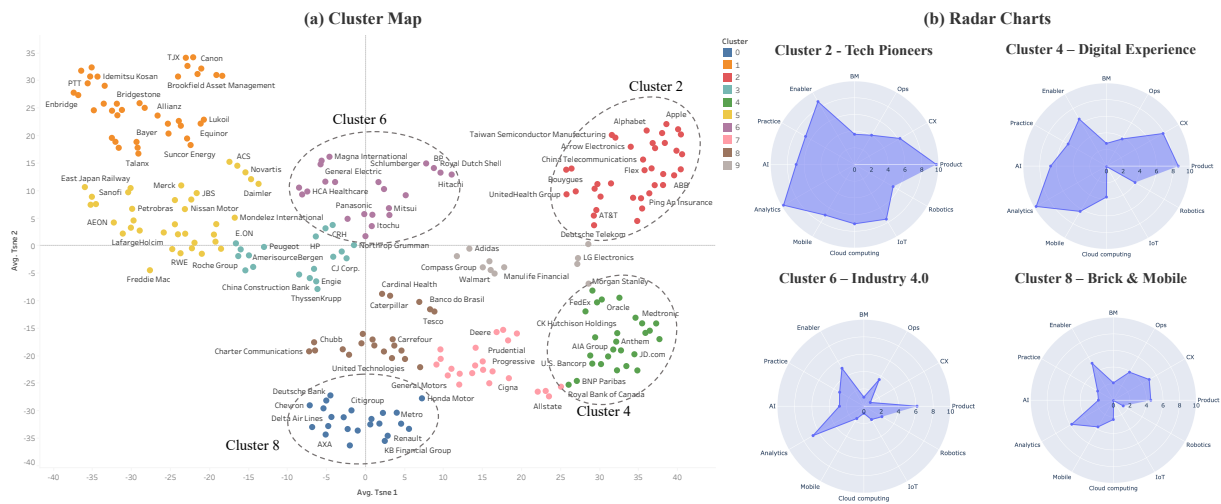

Figure 1: illustration of the digital strategy clusters
(a) Cluster map of the 295 companies based on digital strategy similarity (b) Radar chart of the main 4 archetypes

---

[5] t-SNE hyperparameters were set as (n dimensions=2, random state=42, perplexity=10, learning rate = 200, iterations=1000)

# 6 Conclusion

In this study, we present a deep learning framework for measuring the digital strategy of companies by using earnings calls transcripts. In the process, we demonstrate the practical value of state-of-the-art NLP models beyond inference. The results from our experiment enhance the understanding of digital strategy by identifying four distinct digital strategy archetypes. Our findings may serve as a baseline for companies attempting to formulate or benchmark their digital strategy as well as an empirical foundation for further research. For future studies, we recommend investigating the causal relationship between aspect maturity and financial performance.

## Acknowledgements

We thank the anonymous reviewers for their helpful feedback. We also thank Tim Pearce, Farah Shamout, and Adrià Salvador Palau for their insightful and valuable feedback.

# Appendix A. List of Keyword Terms

This list presents 17 topics of interest detailed by 350 definitional terms. This list of keyword terms was used to preselect relevant sentences for analysis.

| Topic | Keyword | Keyword terms |
|---|---|---|
| Business Value | Digital Product | smart product*, connected product*, software as a service, Saas, platform as a service, Paas, platform, product as a service |
| | Digital Customer Experience | digital experience, customer experience, CX, user experience, UX, user journey, customer journey, digital engagement, customer engagement, personalization, personalisation, digital marketing, recommendation, market place, marketplace, e-commerce platform, digital service, digital services, e-service, online chat, chatbot, \bapp\b, digital chnnel, omnichannel |
| | Digital Operations | process automation, process mining, process analytics, process optimization, efficiency, cost saving, cost reduction, reduce cost, reducing cost, automation, predictive maintenance, ERP, supply chain, logistics, operations |
| | Digital Business Model | business model, new market, new segment, monitization, Saas, software as a service, on-demand, product as a service, value proposition, freemium, subscription, marketplace, ad-revenue, ads, peer-to-peer, two-sided, double-sided |
| Strategy Management | Enablers | digital strategy, digital business strategy, digital transformation strategy, governance, priositization, prioritization, digital vision, digital leadership, leadership support, leadership buy-in, communication, digital goals, data scientist, data analyst machine learning engineer, developer, coder, programmer, chief digital officer, CDO, head of digital transformation, head of digital, product manager, product owner, cross-functional, scrum master, agile coach, innovation manager, Data lake, data warehouse, middle ware, enterprise architecture, digital tools, digital workplace, digital integration, chat, video call, CRM, ERP, service oriented architecture, \bSOA\b |
| | Practices | Agile, scrum, MVP, minimum viable product, sprint, design thinking, business experiment, DevOps, \bepic\b, feature, user story, product owner, product manager, collaboration, cross functional, cross-functional, A/B testing, exploratory data analysis, data analysis, decision support system, dashboard, hypothesis testing, experimental design, product metrics, user metrics, usage metrics, click through rate, conversion rate, click stream, digital marketing, customer segmentation, risk modelling, simulation, decision analytics, decision support system, project management, digital skills, digital leadership, transformation, data analysis, social media management, social listening, user research, UX research, UX design, UI design, programming, coding, lean startup, experimentation, incubator, accelerator, innovation lab, digital lab, digital transformation, open innovation, design thinking, design sprint, digitalization, digitalisation, digitization, digital technolog[a-z]* |

| Topic | Keyword | Keyword terms |
|---|---|---|
| Digital Technology | AI | \bAI\b, artificial intelligence, NLP, natural language processing, natural language understanding, NLU, natural language generation, NLG, speech recognition, sentiment analysis, speech to text, text to speech, deep learning, machine learning, \bML\b, neural network, algorithm, generative adversarial network, GANs, supervised learning, unsupervised learning, reinforcement learning, semi-supervised learning, active learning, self learning, transfer learning, back propogation, tensorflow, Salesforce Einstein, IBM Watson, kaggle, AI as a service, Microsoft azure ML, AutoML, autonomous vehicles, computer vision, image recognition, pattern recognition, cognitive computing, predictive analytics, predictive maintenance, algorithmic trading, clustering, dimensionality reduction, t-sne, PCA, principal component analysis, chatbot, \bbot\b, RPA, robotic process automation, matrix factorization, collaborative filtering, recommender system, recommendation engine, graph mining, graph theory, cortana, alexa, google assistant |
| | Cloud computing | cloud computing, cloud native, cloudless, distributed cloud, distributed computing, clustered computing, hybrid cloud, platform as a service, edge computing, cloud api, google cloud, azure cloud, aws cloud, software as a service, cloud applications, cloud, GPU, HPC management, cloud storage, elasticity, elastic computing, the cloud, data platform |
| | IoT | Internet of things, \bIoT\b, industrial internet, IIoT, embedded device, embedded sensor, digital twin, digital thread, building information modelling, BIM, connected devices, connected sensors, IoE, internet of everything, smart machines, connected machines, wearable, cyber physical systems, machine to machine, connected factory, model based definition |
| | Virtual reality | \bVR\b, virtual reality, immersive technologies, mixed reality |
| | Augmented reality | \bAR\b, augmented reality, immersive technologies, mixed reality |
| | Robotics | robots, humanoid, drone, drones, smart robots, smart warehouse, smart spaces, Lidar, computer vision, UAV, autonomous vehicles, swarm robots, industrial robot, robotics, automation |
| | Analytics | analytics, business intelligence, optimization, exploratory data analysis, data science, augmented analytics, descriptive analytics, descriptive statistics, prescriptive analytics, predictive, inference, inferential, customer segmentation, correlation, data visualization, data storytelling, text analytics, data lake, data warehouse, big data, social analytics, , network analytics, network mining, network analysis |
| | Mobile | mobile, smart phone, mobile app, mobile application, mobile platform, mobile solution, mobile technology |
| | Social | social media, social network, content marketing |
| | 3D printing | 3D print[a-z]*, additive manufacturing, 3D scan, material jetting, stereolithography, bioprint, bioprinted organ, Fused deposition modeling, Digital Light Processing, Selective Laser Sintering, Selective laser melting, Laminated object manufacturing, Digital Beam Melting |
| | Blockchain | blockchain, distributed ledger, decentralized, smart contracts, cryptocurrency, \bICO\b, initial coin offering, asset tokenization |

## Appendix B. Definition & Frequency of Labels

This list provides a definition of the labels and their frequency in the whole dataset (training and inference). The definitions were used to guide the annotation process. All labelling was carried out by the authors due to the need for domain knowledge in relation to digital strategy.

| Group | Aspect | Keyword definition | Count |
|---|---|---|---|
| Business Value | Digital Product | Enhancing product value through digital features usually requires integrating a combination of products, services and data | 17,936 |
| | Digital Customer Experience | To create a seamless, omnichannel experience that makes it easy for customers to order, inquire, pay and receive support in a consistent way from any channel at any time | 5,107 |
| | Digital Operations | The technology and business capabilities that ensure the efficiency, scalability, reliability, quality and predictability of core operations | 3,399 |
| | Digital Business Model | A significantly new way of creating and capturing business value that is embodied in or enabled by digital technologies | 1,105 |
| Strategy Management | Enablers | All measures that a company have in place to become digitally enabled such as digital strategy, technological infrastructure, and governance | 11,035 |
| | Practices | All activities that reflects digital maturity such as agile practices, data-driven decision making, and digital innovation | 2,245 |
| Digital Technology | AI | Only the explicit mention of digital technologies using commercial or technical terms was used in labelling and inference. Therefore, no definition was needed for this task. | 2,897 |
| | Cloud computing | | 3,931 |
| | IoT | | 1,565 |
| | Virtual reality | | 290 |
| | Augmented reality | | 223 |
| | Robotics | | 1,509 |
| | Analytics | | 2,916 |
| | Mobile | | 6,502 |
| | Social | | 773 |
| | 3D printing | | 26 |
| | Blockchain | | 413 |

| Maturity | Definition | Count |
|---|---|---|
| Plan | Digital initiative is or being planned | 4,286 |
| Pilot | Digital initiative is being developed or piloted | 14,220 |
| Release | Digital initiative is launched and making active contribution to the business | 12,115 |
| Advance | Digital initiative is being pioneered and making significant business impact | 1,3754 |